# Self-Supervision in Time for Satellite Images(S3-TSS): A novel method of SSL technique in Satellite images

Akansh Maurya
akma00001@stud.uni-saarland.de

Hewan Shrestha
hesh00001@stud.uni-saarland.de

Mohammad Munem Shahriar
mosh00001@stud.uni-saarland.de

**Affiliated University:** Universität des Saarlandes

## Abstract

*With the limited availability of labeled data with various atmospheric conditions in remote sensing images, it seems useful to work with self-supervised algorithms. Few pretext based algorithms including from rotation, spatial context and jigsaw puzzles are not appropriate for satellite images [8]. Often, satellite images have a higher temporal frequency. So, the temporal dimension of remote sensing data provides natural augmentation without requiring us to create artificial augmentation of images. Here, we propose S3-TSS, a novel method of self-supervised learning technique that leverages natural augmentation occurring in temporal dimension. We compare our results with current state-of-the-art methods and also perform various experiments. We observed that our method was able to perform better than baseline SeCo [7] in four downstream datasets. Code for our work can be found here: https://github.com/hewanshrestha/Why-Self-Supervision-in-Time*

## 1. Introduction

In recent years, machine learning has made remarkable advances, fueled by a wide range of learning paradigms. The evolution of machine learning has led us to deep learning. With the help of deep learning, we were able to unfold the mystery behind the applications like image classification, object segmentation, semantic segmentation and many more. Among the learning methods, the core methods that support the development of artificial intelligence include supervised, unsupervised, and self-supervised learning. To forecast outcomes and make wise judgments, supervised learning involves using labeled data to train machine learning models. Contrarily, unsupervised learning concentrates on finding underlying structures, relationships, and patterns in unlabeled data, frequently yielding insights that would be difficult to come across otherwise. Self-supervised learning, a new paradigm, has drawn a lot of interest since it takes advantage of algorithms' intrinsic capacity to produce supervisory signals from unlabeled data.

The effectiveness of Deep Learning techniques is highly dependent on the quantity and precision of training data. A significant dataset requires challenging work, a lot of time and money, and there is a chance that human error will occur. Numerous fields, including medicine, satellite imaging, and surveillance film, have access to vast amounts of unlabeled data. Self-supervised learning strategies seek to use these underutilized resources for training in order to solve this.

Satellite images undergo natural transformations over time, including stationary alterations such as lightning, solar radiation, weather conditions, and day-night transitions, influenced by factors like fog and clouds. These images also capture stationary elements like buildings and trees, subject to seasonal changes. Moreover, non-stationary modifications involve dynamic elements such as moving cars and ongoing construction activities. The complexity of these changes defies replication through artificial augmentation techniques. **In our work, we endeavor to explore the potential of harnessing the inherent Natural Augmentation occurring within satellite images over time through a self-supervised learning approach.**

## 2. Related Work

There has been several works done in accordance with self-supervised learning. In particularly, some studies have been done upon remote sensing data with the help of self-supervised models. In one of those paper [11], the researchers proposed a multi-task framework to simultaneously learn from rotation pretext and scene classification to distill task-specific features adopting a semi-supervised perspective. For applications like change detection or crop type classification, the temporal stamps of remote sensing data are crucial. Four different types of datasets have been used to classify remote sensing scenes. In their proposed framework, they have achieved learning of different discriminative features without any overheard parameters.

In case of satellite images, some traditional changes take place depending on the time. It becomes harder to detect and annotate these changes manually. To find a solution, Dong et al. [3] proposed a self-supervised representation learning technique for change detection in distant sensing after quantifying temporal context by coherence in time. In this study, their proposed algorithm can be able to transform images from two satellites into getting more precise representation of a feature without any additional overheads with the help of self-supervision.

Mañas et al. [7] proposed Seasonal Contrast (SeCo) approach which involves compiling large datasets of unlabeled, uncurated satellite photos and using a self-supervised learning technique for pre-training remote sensing representations. The researchers on this paper have discovered the natural augmentations that took place on the satellite images in the SeCo dataset.

On another paper [8], empirical results show that models trained with SeCo dataset outperform ImageNet pre-trained models and state-of-the-art self-supervised learning methods on various tasks. This motivates us more to include the SeCo dataset into our proposed methods.

## 3. Method

### 3.1. Model Training Architecture

In this section, we present our methodology for self-supervised learning of satellite images. Our method, called S3-TSS: Self-Supervision in Time for Satellite Images, is inspired by DINO [1], a state-of-the-art self-supervised learning method that uses vision transformers. DINO works by training a student network to predict the output of a teacher network, which is updated with a momentum encoder. The student and teacher networks are trained with a cross-entropy loss, without any contrastive or clustering terms. Global and local crops are different image patches that are used as inputs for the student and teacher networks in DINO. Global crops have a resolution of 224 x 224 pixels and are randomly resized and cropped from the original image. Local crops have a resolution of 96 x 96 pixels and are randomly resized and cropped from a smaller region of the original image. The global and local crops are used to create different views of the same image, which are then aligned by the cross-entropy loss between the student and teacher outputs. This way, DINO can learn to extract features that are invariant to different scales and regions of the image.

In our work, we limited our studies to the ResNet-18 [5] architecture as a backbone. Instead of using artificial augmentations, we used time as a natural augmentation. For one particular geolocation, we had five images in time. We then randomly generated 30 local crops and 10 global crops from these five images. We followed the same cropping ratio of 224 x 224 pixels for global crops and 96 x 96 pixels for local crops as presented in the DINO paper. Our goal was to learn a student model that was better than the teacher model and hence more numbers and difficult augmentations, in this case, crops, were given as input to the student model and global crops acted as input to the teacher model. As mentioned earlier, we used ResNet-18 as a backbone followed by an MLP-based projection head. This MLP took input of 512 features from the backbone and was followed by 3 layers with 512, 64, and 2048 neurons in each. In the teacher's head, an extra centering operation was performed before softmax, which helped in preventing collapse of the model. The weights of the teacher model while training were updated by exponential moving average from the student model. This also helped in avoiding collapse. The overview of the method is shown in diagram 1. After training the model, we discarded the projection head and used the teacher model backbone on downstream tasks which are described in the following sections.

### 3.2. Pre-training Dataset for Self-Supervised Learning

For self-supervised learning algorithms, we use some pretraining and test on downstream tasks of interest. In our project, we have used the Seasonal Contrast (SeCo) dataset [7] for our pre-training task. The SeCo dataset is a remote sensing dataset created from Sentinel-2 [4] tiles without manual human annotation. SeCo dataset has a total of 100,000 images with each image having 5 different seasonal variants in time.

### 3.3. Datasets for Downstream Tasks of Interest

For our downstream tasks of interest, we have used four different remote sensing datasets namely: EuroSAT Dataset [6], Aerial Image Dataset(AID) [9], UCMerced Land Use Dataset [10] and WHU-RS19 Dataset [2].

#### 3.3.1 EuroSAT Dataset

EuroSAT dataset is a remote sensing dataset that covers 13 spectral bands and consists of a total of 27000 labeled and geo-referenced images into ten different classes: Annual-Crop, Forest, HerbaceousVegetation, Highway, Industrial,

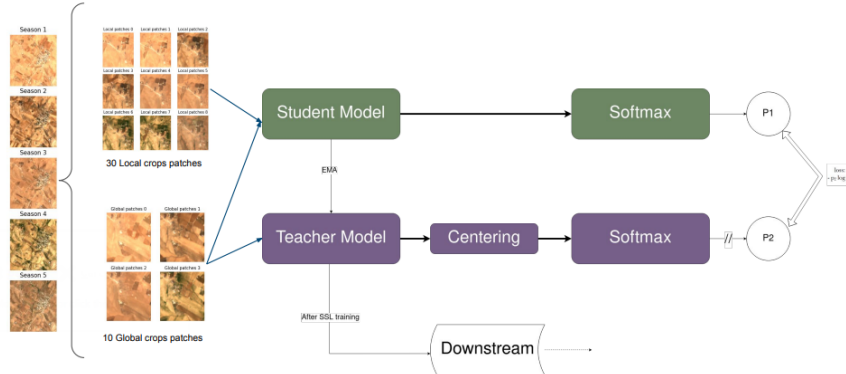

Figure 1. Overview of S3-TSS: Self-Supervision in time for Satellite Images.

Pasture, PermanentCrop, Residential, River, and SeaLake.

### 3.3.2 Aerial Image Dataset (AID)

AID is a large-scale aerial image dataset, which is made up of 10000 images distributed into the following 30 aerial scene types: airport, bare land, baseball field, beach, bridge, center, church, commercial, dense residential, desert, farmland, forest, industrial, meadow, medium residential, mountain, park, parking, playground, pond, port, railway station, resort, river, school, sparse residential, square, stadium, storage tanks, and viaduct.

### 3.3.3 UCMerced Land Use Dataset

The UCMerced Land Use Dataset contains 100 images for each of the following 21 classes: agricultural, airplane, baseball diamond, beach, buildings, chaparral, dense residential, forest, freeway, golf course, harbor, intersection, medium residential, mobile homepark, overpass, parking lot, river, runway, sparse residential, storage tanks, and tennis court. Each image in this dataset measures 256x256 pixels.

### 3.3.4 WHU-RS19 Dataset

WHU-RS19 Dataset is a set of satellite images that provides high-resolution remote sensing images up to 50cm. There are a total of 19 classes of scenes, including airport, beach, bridge, commercial, desert, farmland, football field, forest, industrial, meadow, mountain, park, parking, pond, port, railway station, residential, river, and viaduct, with about 50 samples of high-resolution imagery in each class.

### 3.4. Training hyperparameters and evaluation metric

To ensure an equitable comparison of the results obtained through the S3-TSS approach with those generated by alternative methods, we maintained consistency in all hyperparameters. The specific hyperparameter settings utilized during the training of the ResNet-18 model for the downstream task are elucidated in Table 1. Furthermore, to validate the applicability of our approach across diverse scenarios, we assessed its performance across the four datasets previously described. Notably, for the sake of conducting a comparative evaluation, adjustments were exclusively made to the pretrained model weights and the model's initialization procedure. We used Linear probing and finetuning as metric. Linear probing assesses the quality of learned representations by training a linear classifier on fixed features, while fine-tuning adapts the entire model to a downstream task, reflecting the transferability of acquired features.

| Pre-training | Downstream Dataset | Hyperparameters | Metric |
|---|---|---|---|
| Random Initialization | EuroSAT | Optimizer: Adam | |
| ImageNet Initialization | AID | Learning Rate: 0.001 | Linear-probing |
| SeCo Initialization | UCMerced Land Use | Epochs: 20 | Fine-tuning |
| S3-TSS [Ours] | WHU-RS19 | Batch size: 64 | |
| DINO with artificial augmentation | | StepLR scheduler | |

Table 1. Hyperparameter selection

## 4. Experimental Results and Analysis

### 4.1. Experiment 1

We conducted a series of four experiments to address the overarching inquiry: Can Natural Augmentation yield superior performance compared to Artificial Augmentations? Experiment 1 encompassed a self-supervised framework comprising two primary stages. The initial phase involved training a backbone model with a pretext task, utilizing either absent or fabricated labels. Subsequently, in the second phase, the identical backbone model was employed for supervised training. Given the challenge of establishing a suitable termination criterion in the absence of labels during the initial step, Experiment 1 was dedicated to elucidating and solidifying the termination criteria for our proposed approach. Specifically, a subset of 20,000 images was extracted from a pool of 100,000 images in the SeCo dataset. Employing the DINO method, the model was trained over 30 and 100 epochs, noteworthy is the utilization of artificial augmentations for this stage. Post-training, the backbone model was further fine-tuned across four distinct datasets, with variations introduced in terms of the proportion of data subjected to supervised learning. As illustrated in Figure 2, the outcomes pertaining to the EuroSat dataset are presented; comprehensive outcomes for the remaining datasets are accessible via the provided link. Discerning from Figures **??** and **??**, it can be inferred that an extended number of training epochs yields enhanced performance. This assertion is supported by the observation that the green curve, representing the model trained over 100 epochs, consistently outperforms its counterparts. Intriguingly, the linear probing results indicated that the DINO-initialized model in both training configurations surpassed the performance of models initialized with ImageNet pretraining. Guided by these findings, subsequent experiments will adopt the criterion of 100 epochs as the established termination point.

### 4.2. Experiment 2

Experiment 2 was undertaken with the principal aim of comprehending the nuanced impact of the quantity of unsupervised data deployed during the pretraining phase. This investigation unfolded across two distinct scenarios: one involving a dataset of 20,000 images and the other comprising 100,000 images. All other experimental parameters were held constant, paralleling the conditions established in Experiment 1. The findings gleaned from Figure 3 substantiate that the red curve, corresponding to the model pretrained on the larger corpus of 100,000 SeCo dataset images, emerges as the dominant performer. Particularly noteworthy is the outcome of linear probing illustrated in Figure **??**, where the model's performance outpaces ImageNet by a substantial margin. Additionally, the performance parity observed in the fine-tuning context depicted in Figure **??**, despite ImageNet's significantly larger pool of 1 million images compared to SeCo's 100,000, is a notable observation. Guided by these insights, forthcoming experiments shall adopt the 100,000-image configuration for the pretraining phase. Results for the remaining datasets are accessible via the provided link.

### 4.3. Experiment 3

In the context of Experiment 3, we introduced our Self-Supervision in Time for Satellite Images (S3-TSS) methodology, designed to facilitate a comparative assessment against the current state-of-the-art SeCo approach as expounded in reference [7], as well as DINO when employed in conjunction with artificial augmentations. The procedural intricacies of the S3-TSS pretraining method are elaborated upon within the methodology section. The outcomes of Experiment 3 are presented through Figure 4, focusing on the EuroSAT dataset. Specifically, within Figure **??**, the purplish curve signifies the outcomes derived from the S3-TSS approach. The observations therein demonstrate the comparable performance of natural augmentation in S3-TSS, exhibiting results on par with the artificial augmentation strategies adopted in both DINO and SeCo. Subsequently, in Figure 4, S3-TSS exhibits superior performance compared to the SeCo initialization. Notably, the adoption of natural augmentation within S3-TSS confers the advantage of temporal efficiency, as it obviates the need for resource-intensive artificial augmentation endeavors. It is prudent to acknowledge that while DINO, in tandem with artificial augmentations, achieves a performance edge over S3-TSS, this does come at the expense of heightened computational demands. Results for the remaining datasets are accessible via the provided link.

## 5. Conclusion

The empirical investigations undertaken in this study provide valuable insights into the dynamics of self-supervised learning methodologies, particularly with respect to training duration, dataset scale, and the efficacy of a novel approach termed S3-TSS. The subsequent analysis yields noteworthy observations in the context of fine-tuning and linear-probing evaluations.

- **Experiment 1**: Empirical evidence substantiates the notion that the augmentation of self-supervised training epochs correlates positively with enhanced model performance.

- **Experiment 2**: The discernment of the pivotal role played by the quantity of data utilized for self-supervised learning underscores the significance of transitioning from a 20,000-image dataset to a more expansive pool of 100,000 images, resulting in considerable performance gains.

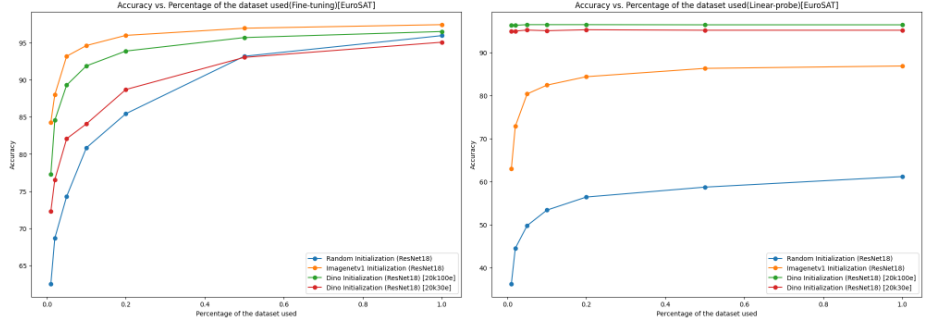

Figure 2. Fine-tuning and linear-probing results for Experiment 1

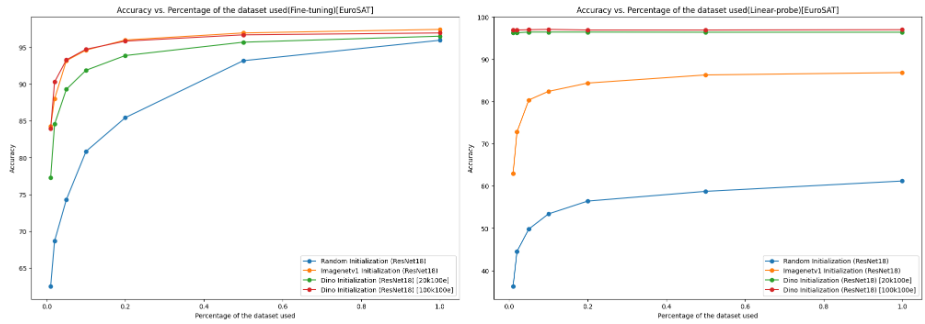

Figure 3. Fine-tuning and linear-probing results for Experiment 2

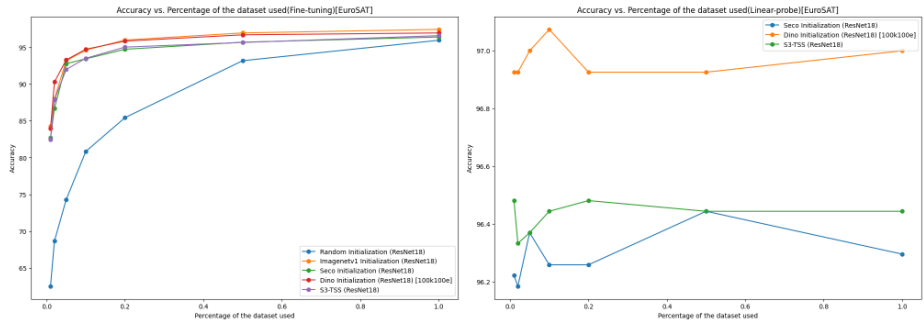

Figure 4. Fine-tuning and linear-probing results for Experiment 3

- **Experiment 3**: The introduction of the SeSelf-Supervision in Time for Satellite Images (S3-TSS) method yields commendable results, surpassing the SeCo baseline without resorting to artificial augmentation. However, it is noteworthy that the DINO self-supervised learning (SSL) approach, coupled with artificial augmentation, demonstrates superior performance albeit at a heightened computational cost.

- **Fine-tuning Analysis**: The trend consistently observed across various datasets indicates that, in fine-tuning scenarios, models initialized with ImageNet consistently outperform other initialization methods. It is pertinent to acknowledge that the notable performance discrepancy is influenced by the significant dataset scale disparity, with ImageNet comprising 1 million images in contrast to the study's dataset encompassing 100,000 images.

- **Linear-Probing Evaluation**: Both the S3-TSS and DINO initializations outshine the performance of models initialized with ImageNet across the linear-probing evaluation, further highlighting the efficacy of these methods in capturing salient feature representations.

We also performed some more experiments with ResNet-50 Architecture, please refer to this link.